\documentclass[letterpaper, 10 pt, conference]{ieeeconf}   

\IEEEoverridecommandlockouts                              

\usepackage{array} 
\usepackage{graphicx}
\usepackage{amsmath} 
\usepackage{comment}
\usepackage{booktabs}       
\usepackage{multirow}
\usepackage{makecell}
\usepackage{adjustbox}
\usepackage{xcolor}
\usepackage{cite}

\usepackage{bbold}
\usepackage{amsfonts}
\usepackage{amssymb}
\usepackage{txfonts}
\usepackage{pifont}

\usepackage{booktabs}
\usepackage{nicematrix}

\newcolumntype{C}{>{\centering\arraybackslash}p{4.5em}}

\newcommand{\cmark}{\ding{51}}%

\title{\LARGE \bf
HAPFI: History-Aware Planning based on Fused Information
}

\author{Sujin Jeon$^{*}$ $\,$ Suyeon Shin$^{*}$ $\,$ Byoung-Tak Zhang$^{1}$% <-this % stops a space
\thanks{$^*$Authors have equal contributions}
\thanks{$^{1}$AI Institute, Seoul National University}
\thanks{All authors are affiliated with Interdisciplinary Program in AI (IPAI), Seoul National University}
}

\begin{document}

\maketitle
\thispagestyle{empty}
\pagestyle{empty}

%%%%%%%%%%%%%%%%%%%%%%%%%%%%%%%%%%%%%%%%%%%%%%%%%%%%%%%%%%%%%%%%%%%%%%%%%%%%%%%%
\begin{abstract}
Embodied Instruction Following (EIF) is a task of planning a long sequence of sub-goals given high-level natural language instructions, such as ``\textit{Rinse a slice of lettuce and place on the white table next to the fork}''. 
To successfully execute these long-term horizon tasks, we argue that an agent must consider its past, i.e., historical data, when making decisions in each step.
Nevertheless, recent approaches in EIF often neglects the knowledge from historical data and also do not effectively utilize information across the modalities.
To this end, we propose History-Aware Planning based on Fused Information(HAPFI), effectively leveraging the historical data from diverse modalities that agents collect while interacting with the environment.
Specifically, HAPFI integrates multiple modalities, including historical RGB observations, bounding boxes, sub-goals, and high-level instructions, by effectively fusing modalities via our Mutually Attentive Fusion method.
Through experiments with diverse comparisons, we show that an agent utilizing historical multi-modal information surpasses all the compared methods that neglect the historical data in terms of action planning capability, enabling the generation of well-informed action plans for the next step.
Moreover, we provided qualitative evidence highlighting the significance of leveraging historical multi-modal data, particularly in scenarios where the agent encounters intermediate failures, showcasing its robust re-planning capabilities.
\end{abstract}
%%%%%%%%%%%%%%%%%%%%%%%%%%%%%%%%%%%%%%%%%%%%%%%%%%%%%%%%%%%%%%%%%%%%%%%%%%%%%%%%
\section{INTRODUCTION}

\begin{figure}[t]
\centering
\includegraphics[width=\linewidth]{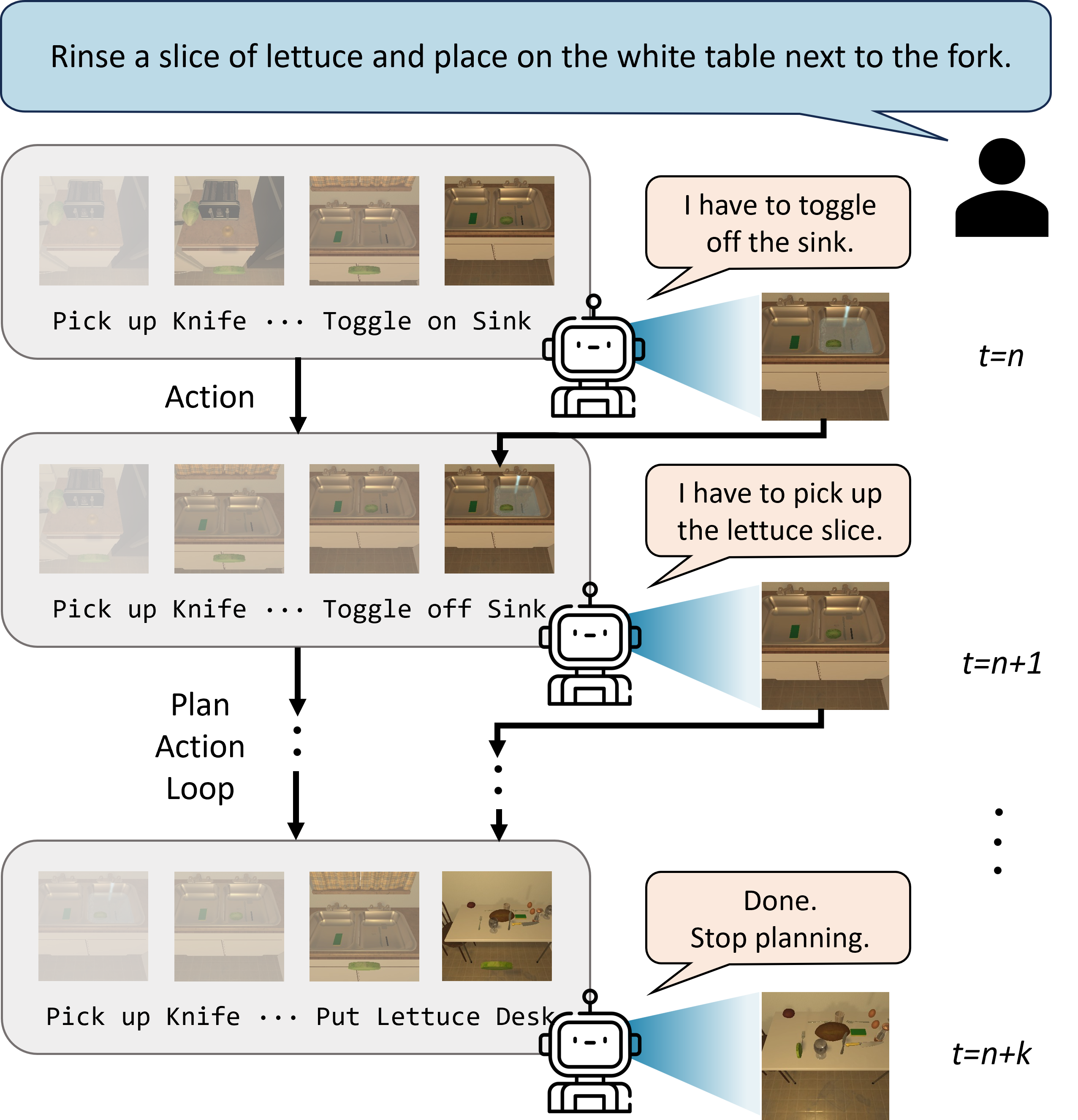}
\caption{\textbf{Planning Procedure of the EIF Task Employing HAPFI.} The agent utilizes visual and sub-goal history as demonstrated in the gray box. It determines its next action by referencing this historical information, the current RGB observation, and the provided high-level instruction. This decision-making process is repeated in a continuous plan-action loop until the agent decides the completion of the entire task.}
\label{fig:eif_overview}
\end{figure}

The substantial progress in artificial intelligence has heightened expectations for embodied agents capable of interacting with real-world environments and executing interactive actions.
Consequently, ongoing research has been focusing on the development of embodied agents, including robots, with the capacity to emulate human abilities in efficiently processing multifaceted, long-term information.
One notable task of this is Visual-Language Navigation (VLN), where the agents are directed to specific destinations using natural language instructions. 
Recent attention has been on tackling the more complicated problem, Embodied Instruction Following (EIF), which aims to reflect real-world complexity better.
For example, in Figure \ref{fig:eif_overview}, an agent has to execute the instruction \textit{``Rinse a slice of lettuce and place on the white table next to the fork''}, which has a complexity level similar to real-world tasks.
This task is more challenging due to the diversity of action spaces and the need to predict a sequence of sub-goals with a long-tern horizon. 
      
As research in the field of EIF progresses, a variety of approaches have emerged for predicting interactive sub-goals, including actions, objects, and receptacles, which the agent executes on the environment \cite{ABT,ALFRED,CAPEAM,Episodic_transformer,FILM,HITUT,HLSM,LGS-RPA,LLM-Planner,LWIT,MOCA,Prompter}.
The first problem in this research progress has been the agent's under-utilization of multi-modal information. In the past approaches, there have been works \cite{FILM,Prompter,CAPEAM} that use only natural language instruction. However, on account of the complexity of scenarios, some approaches incorporate modalities, such as RGB observations \cite{MOCA,Episodic_transformer,HITUT}, multiple surrounding views of the scene \cite{LWIT, ABT}, or semantic information \cite{HLSM,LGS-RPA} with natural language, to gain a deeper understanding of intricate scenarios. The point we have focused on is that the methods employed to integrate these diverse modalities have some limitations. These approaches depend on the simple concatenation of information from each modality and feeding it into a Transformer-based encoder, RNN-based decoder, or Multi-layer Perceptrons (MLP) for integration. 
According to significant research in the field of visual grounding \cite{Uniter,MDETR,Unicoder,Visualbert,Oscar,Vilbert,Vlbert,Lxmert,Unified_VLP,GLIP}, effectively fusing visual and language features is of paramount importance.
Inspired by prominent research \cite{GLIP} in the visual grounding task, we adopt an early and deep fusion approach to overcome the limitations of previous methods, which employed late and weak feature integration methods. Our approach aims to enhance the integration of various modalities to solve agent's under-utilization of multi-modal information.

The second problem we must address is that agents need to comprehend the scenario leading up to the current event, as they need to comprehend the long temporal context. 

Current states alone may not encompass sufficient information, necessitating the incorporation of historical data.
Most existing methods either do not use historical information \cite{FILM, Prompter, ALFRED, CAPEAM, ABT, MOCA, LWIT } or rely on only sub-goal history \cite{LGS-RPA,HLSM,HITUT} for next-step planning, except for a work \cite{Episodic_transformer} that utilizes both vision and sub-goal history. Despite this fact, due to its weak integration mechanism, as highlighted above, there are limitations in performance improvement; even step-by-step instructions are provided.

Our approach, \textbf{H}istory-\textbf{A}ware \textbf{P}lanning based on \textbf{F}used \textbf{I}nformation (HAPFI), focuses not only on the utilization of historical context to temporal integration, but also fusion approach for integrating them.
Specifically, HAPFI aims to predict the requisite sub-goals for executing high-level instructions effectively. 
HAPFI comprises three pivotal components: (1) Historical Multi-modal Feature Integration: Information from various modalities, along with its historical context, including RGB observations, bounding boxes, sub-goals, and natural language instructions, are initially encoded. Then, visual and linguistic features are subsequently integrated, respectively. (2) Mutually Attentive Fusion: HAPFI incorporates a feature integration layer that makes deep fusion from diverse modalities, making them attend to each other. (3) Sub-goal Classification: The model predicts requisite actions, objects, and receptacles through task-specific heads. Then, an agent iteratively updates the consequences of actions through interactions with the environment until it successfully completes the high-level instruction. 

Through experiments with diverse comparisons, we show that HAPFI utilizing historical multi-modal information surpasses all the compared methods that neglect the historical data in terms of action planning capability. HAPFI enables an agent to generate well-informed action plans for the next step, thus enhancing their ability to interact within their environment.
Moreover, we provide qualitative evidence highlighting the significance of leveraging historical data, particularly in scenarios where the agent encounters intermediate failures, showcasing its robust re-planning capabilities.
 
In summary, our contributions are twofold:        
\begin{itemize}
\item We propose a novel framework HAPFI, that addresses action planning in the context of Embodied Instruction Following (EIF) by deeply fusing the features of historical multi-modalities.
\item We demonstrate the planning ability of HAPFI through sub-goal planning experiments followed by the qualitative analyses on the ALFRED benchmark.
\end{itemize}

\begin{figure*}[t]
\centering
\includegraphics[width=\linewidth]{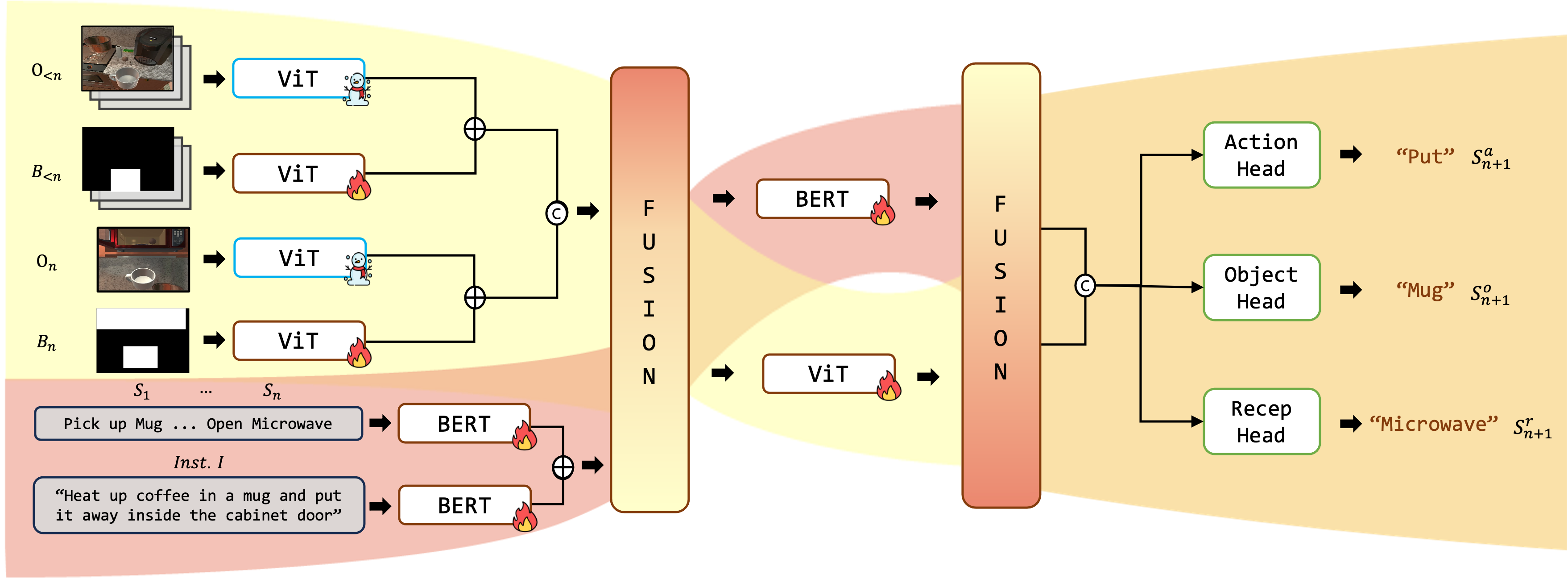}
\caption{\textbf{Overview of History-aware planning based on Fusion Integration} Image history $O_{<n}$ and current image $O_{n}$ are encoded through a pretrained Vision Transformer (ViT) while bounding bbox (bbox) $B_{<n}$ history and current bbox $B_{n}$ are encoded using ViT trained from scratch. Sub-goal history $S_{<n}$ and high-level instruction $I$ are encoded by BERT architecture trained from scratch. The information encoded in each modality is integrated through the fusion module and then predicts actions $S^a_{n}$, objects $S^o_{n}$, and receptacles $S^r_{n}$ separately through task-specific heads. c denotes concatenation, and + denotes element-wise sum.}
\label{fig:framework}
\end{figure*}

%%%%%%%%%%%%%%%%%%%%%%%%%%%%%%%%%%%%%%%%%%%%%%%%%%%%%%%%%%%%%%%%%%%%%%%%%%%%%%%%
\section{RELATED WORK}
Robotics has recently prioritized embodied agents that can integrate visual and linguistic data to interact with real-world environments, resulting in the progress of Visual-Language Navigation (VLN) \cite{VLN_look,VLN_speaker,VLN_tactical,VLN_regretful,VLN_reinforced,VLN_interpreting,VLN_selfsupervised} and Embodied Instruction Following (EIF) \cite{HITUT,HLSM,FILM,CAPEAM,LGS-RPA,LLM-Planner,Prompter,ABT,LWIT,MOCA}.
Here, we focus on planning actions from the latter line of work, EIF, that aims to integrate both navigation and interaction.

One noteworthy benchmark in this regard is ALFRED \cite{ALFRED}. Unlike earlier benchmarks \cite{pre_alfred_1,pre_alfred_2,pre_alfred_3} marked by confined environments and limited object variety, ALFRED seeks to emulate real-world situations, infusing tasks with increased complexity and diversity.

In the early stages of research, there was a significant focus on studies in which agents executed actions by following step-by-step instructions to accomplish high-level goals \cite{ALFRED,ABT,LWIT,MOCA,Episodic_transformer}.
However, recent methods have introduced approaches that provide only high-level goals rather than detailed step-by-step guidance \cite{HITUT,HLSM,LGS-RPA,FILM,CAPEAM,Prompter,LLM-Planner}. 
These under-specified instructions referred to as high-level instructions, are insufficient for task completion, which demands a profound understanding of multifaceted temporal context.
We intend to focus on the ability to plan sub-goals relying on high-level instructions, without step-by-step instructions.

When considering existing action planning approaches from a modality perspective, it's worth noting that some of these methods rely on single-modal information \cite{FILM,CAPEAM}. In most cases, the primary focus is natural language instructions. Even when other modalities (typically images or semantic information) are integrated into predictions, the integration process tends to be relatively weak \cite{HITUT,HLSM,LGS-RPA}. These modalities are processed individually through separate encoders, with integration occurring towards the end. Specifically, features from different modalities, each encoded by distinct encoders, do not mutually attend to each other.

Several approaches have been attempted to address the complexity and long-term horizon in predicting subsequent actions. One of the most recent methods \cite{CAPEAM} involves incorporating context into planning to memorize environmental changes. On the other hand, there have been approaches that focus on incorporating the history of sub-goals \cite{LGS-RPA,HLSM,HITUT} to enhance prediction accuracy.
Most existing methods either do not use historical information for sub-goal planning or rely on sub-goal history only. 

Building upon these insights, our proposed method, HAPFI, incorporates multi-modal historical information related to RGB observations, bounding boxes, predicted sub-goals, and high-level instruction. This integration empowers our model to comprehend intricate scenarios and predict subsequent actions effectively, which proves particularly valuable in tasks characterized by extended time horizons. Furthermore, inspired by the visual grounding method GLIP \cite{GLIP}, we adopt deep and early fusion mechanisms. This mechanism focuses on achieving more accurate action planning by effectively fusing features from diverse modalities.

\section{METHOD}
\label{sec:method}

We present HAPFI (History-Aware Planning based on Fused Information), a model designed to deeply fuse historical multi-modalities, enhancing the efficiency of action planning tasks. HAPFI comprises four essential phases: Historical Visual Feature Integration, Historical Linguistic Feature Integration, Mutually Attentive Fusion, and Sub-goal Classification. 
We proceed by providing an explanation of each phase. 
First, we elucidate the process of encoding and integrating visual features~\ref{ssec:A} and linguistic features~\ref{ssec:B} , respectively. Then, we detail the fusion of historical multi-modalities, emphasizing their mutual attention~\ref{ssec:C}. Finally, we explain HAPFI's subgoal inference mechanism in detail~\ref{ssec:D}.

\subsection{Historical Visual Feature Integration}
\label{ssec:A}

This phase aims to capture a wide range of visual information obtained from the environment, including both bounding box and RGB observation. Initially, we separately encode the bounding box data and RGB observation of the current scene, combining them afterward. Following that, we employ the same process to accumulate historical visual features, subsequently integrating all visual features into a unified representation.

\noindent \textbf{Bounding Box Image.}
By using bounding box coordinates from the detection results, we create a mask that highlights the area within the bounding box. This is achieved by setting the pixels inside the bounding box to the corresponding object class in a zero-valued image. As a result, we obtain a single-channel image where `$0$' represents the absence of an object, and other numbers indicate the presence of a corresponding object. This image is labeled as $B_{n}$, with $n$ meaning the current step.

\noindent \textbf{RGB Image.}
As we predict the agent's actions by a step-by-step procedure, we consider the agent’s viewpoint after the completion of the latest action as the current RGB observation, $O_n$. For instance, if the agent has just picked up an apple, the image of holding the apple within the agent's view is treated as $O_n$.

\noindent \textbf{Visual Modality Information Encoding.}
We employ pretrained ViT\cite{vit} backbone as the RGB observation encoder. Passing $O_n$ through this ViT backbone yields an embedding denoted as $E_n^O$. 
Furthermore, we utilize the ViT trained from scratch as the bounding box encoder for processing $B_n$, resulting in the embedding $E_n^B$.

\noindent \textbf{History Accumulation.}
To ensure memory efficiency, we limit our usage of the historical information to the $l$ most recent data points, resulting in: the bounding box image history, $B_{<n}=[B_{n-l},\cdots,B_{n-1}]$, and the RGB observation history, $O_{<n}=[O_{n-l},\cdots,O_{n-1}]$. To encode $B_{<n}$ and $O_{<n}$, we draw inspiration from the approach used in video-based research, which involves calculating the mean of frames \cite{clip4clip}. In the encoding process for $O_{<n}$, we individually encode each of the $O_k\; (k=n-l,\cdots,n-1)$ using the RGB observation encoder. Subsequently, we compute the mean of the resulting $l$ embedding vectors, $E_k^O$, aligning their size with that of $E_n^O$. Similarly, for $B_{<n}$, we separately encode each of the $B_k\;(k=n-l,\cdots,n-1)$ using the bounding box encoder. We then take the mean of the $l$ resulting embedding vectors, $E_k^B$, making their size consistent with $E_n^B$. 

\noindent \textbf{Feature Integration}
Ultimately, we obtain a visual feature information $V$ that will be utilized in the Mutually Attentive Fusion layer. The procedure can be written as \eqref{eqn:he}, where $+\!\!+$ denotes concatenation.
\begin{equation}\label{eqn:he}
V=\left(\frac{1}{l}\sum_{k=n-l}^{n-1}E_k^O+\frac{1}{l}\sum_{k=n-l}^{n-1}E_k^B\right)+\!\!\!+\,\left(E_n^O+E_n^B\right)
\end{equation}

\subsection{Historical Linguistic Feature Integration}
\label{ssec:B}
In this phase, we leverage two linguistic modalities of information: high-level instruction and sub-goal history. We encode each information separately and integrate the obtained features.

\noindent \textbf{Sub-goal History.}
We consolidate all prior actions, objects, and receptacles into a single string, which serves as our sub-goal history $S_{<n}$. For instance, if the most recent action was `Pickup' and the latest object was `Pencil,' we append `Pickup Pencil' to $S_{<n}$. Each action, object, and receptacle within $S_{<n}$ is separated by a space.

\noindent \textbf{Linguistic Modality Information Encoding and Feature Integration.}
We utilize BERT\cite{bert} as the linguistic modality information encoder. Interestingly, different from RGB observation encoding process, we opted not to use a pre-trained model because experimental results showed superior performance without it. Passing the high-level instruction $I$ into the BERT produces an embedding vector $E^I$. Additionally, by processing the sub-goal history $S_{_<n}$ through the BERT, we obtain another embedding vector $E^S_{<n}$. Lastly, we acquire linguistic feature information $L$ for the Mutually Attentive Fusion layer by simply summing $E^I$ and $E^S_{<n}$. 

\subsection{Mutually Attentive Fusion}
\label{ssec:C}
Inspired by the GLIP\cite{GLIP} framework, we deeply fuse visual feature information $V$ and linguistic feature information $L$ through a multi-head cross-attention mechanism. The process unfolds as follows:
\begin{gather}\label{eqn:eq1}
q=V W_q\;,\; k=L W_k\;; \\ \label{eqn:eq2}
v_V^{}=V W_v^V\;,\; v_L^{}=L W_v^L\;; \\ \label{eqn:eq3}
v^{\ast}_V=\mathrm{softmax}\left(q\cdot k\,/\sqrt{d}\right)v_V^{}\;; \\ \label{eqn:eq4} 
v^{\ast}_L=\mathrm{softmax}\left(q\cdot k\,/\sqrt{d}\right)v_L^{}\;; \\ \label{eqn:eq5}
V_{f_1}=v^{\ast}_L W_p^L+V\;,\;L_{f_1}=v^{\ast}_V W_p^V+L\;; \\ \label{eqn:eq6}
\left(V_{f_2}\,,\,L_{f_2}\right)=\mathrm{Fusion}\left(\mathrm{ViT}\left(V_{f_1}\right)\,,\,\mathrm{BERTLayer}\left(L_{f_1}\right)\right); \\ \label{eqn:eq7} 
F=V_{f_2}\,+\!\!\!+\;L_{f_2}\;. 
\end{gather}
Initially, we create query $q$ and key $k$ pairs by applying query matrix $W_q$ and key matrix $W_k$ to $V$ and $L$ respectively (\ref{eqn:eq1}). Then we get two values $v_V^{}$ and $v_L^{}$ by multiplying different value matrices $W_v^V$ and $W_v^L$ to $V$ and $L$ (\ref{eqn:eq2}). Attention weights are computed using scaled dot-product attention between the queries and keys. These attention weights are transformed into attention scores, which are then used to weight both the $v_V^{}$ and $v_L^{}$. 

The score-weighted linguistic value $v^{\ast}_L$ (\ref{eqn:eq4}) is further processed by a projection matrix $W_p^L$, resulting in the fused embedding. This embedding is added to $V$, generating fused visual feature $V_{f_1}$ (\ref{eqn:eq5}). Likewise, the score-weighted visual value $v^{\ast}_V$ (\ref{eqn:eq3}) is multiplied by another projection matrix $W_p^V$ and added to $L$, producing fused linguistic feature $L_{f_1}$ (\ref{eqn:eq5}). 

$V_{f_1}$ is reshaped into a $14\times14$ tensor and subsequently processed by a ViT, while $L_{f_1}$ is fed into a Bert Layer.
We then perform a second fusion step, resulting in $V_{f_2}$ and $L_{f_2}$, which are subsequently concatenated into $F$. 

\subsection{Sub-goal Classification}
\label{ssec:D}

The concatenated feature $F$ is fed into three separate 3-layer perceptron heads, each responsible for predicting actions, objects, and receptacles.
In action prediction, we classify actions among $7$ interactive actions (`Pick up', `Put', `Open', `Close', `Toggle on', `Toggle off', and `Slice'), `Navigate', and `Stop'. In object prediction, we determine which object the agent should interact with from a pool of $108$ objects. If the action prediction is `Navigate', we predict the destination object. For receptacle prediction, we identify potential receptacles from a set of $38$ receptacles, including 'empty' to signify the absence of a receptacle requirement.

\section{EXPERIMENTS}

\begin{table*}[]
\caption{
\begin{flushleft}
    \textnormal{\textbf{Experimental Results.} 
    The table columns are defined as follows: `Current Step' columns represent the available modalities at the current step, where $O_n$ and $B_n$ denote RGB observation and bounding box data, respectively. `History' columns contain historical modalities, with $O_{<n}$ representing RGB observation history, $B_{<n}$ denoting bounding box data history, and $S_{<n}$ signifying sub-goal history. `Instruction' column only refers to high-level instruction, not step-by-step instruction. Checkmarks indicate that the model in the corresponding row utilizes the modality in the column. In the 'Valid Seen' and 'Valid Unseen' columns, we present the experiment results for each dataset. Bold symbols in numbers denote the highest accuracy, while underlined symbols represent the second-best accuracy. `Total' indicates the accuracy of predicting all the next actions, objects, and receptacles correctly.} 
\end{flushleft}
}

\renewcommand{\arraystretch}{1.5}
\centering
\resizebox{\textwidth}{!}{%
\begin{tabular}{@{}cccccccccccccccc@{}}
\toprule
\multirow{2}{*}{Model} & \multicolumn{2}{c}{Current Step} & \multicolumn{3}{c}{History} & Instruction & \multicolumn{4}{c}{Valid Seen} & \multicolumn{4}{c}{Valid Unseen}  \\ \cmidrule(l){2-3}  \cmidrule(l){4-6} 
 \cmidrule(l){7-7} \cmidrule(l){8-11} \cmidrule(l){12-15}
& $O_{n}^{}$ & $B_n^{}$ & $O_{<n}^{}$ & $B_{<n}^{}$ & $S_{<n}^{}$ & $I$ & Actions & Objects & Receptacles & \textbf{Total} & Actions & Objects & Receptacles & \textbf{Total} \\ \midrule
FILM\cite{FILM}                   
&        &{}     &    &    &  &\cmark  & 94.8    & 88.0    & 92.5        & 82.1  & 95.2    & 86.9    & 92.8        & 80.6  \\
CAPEAM\cite{CAPEAM}                 
&        &{}     &    &    & {}  &\cmark  & -       & -       & -           & 83.1  & -       & -       & -           & 77.3  \\
E.T.\cite{Episodic_transformer}    
& \cmark &{}      & \cmark  &    &{\cmark} &  & 64.6      & 76.4       & -           & 62.6  & 53.8       & 65.9      & -           & 50.9  \\ \midrule
\multirow{4}{*}{Ours}  
&        &{}     &   &    & {\cmark}  & {\cmark}  & 95.9    & 87.1    & 98.8        & 85.4  & 96.5    & 89.2   & \underline{98.8}         & 87.9  \\
& \cmark      &{\cmark}    &   &    & {}   & {\cmark}   & 85.3    & 80.1    & 96.7        & 75.0  & 75.3    & 70.6   & 91.6         & 58.0  \\
& \cmark      &{}     & \cmark &    & {\cmark}    & {\cmark} &  \underline{97.2}       & \underline{88.9}       & \underline{98.9}           & \underline{87.8}     & \underline{97.8}
& \underline{91.5}      & 98.6            & \underline{90.0}     \\

& \cmark      &{\cmark}     & \cmark  & \cmark  & {\cmark} & {\cmark} & \textbf{98.7}    & \textbf{92.1}    & \textbf{99.2}        & \textbf{91.4}  & \textbf{98.4}    & \textbf{92.4}    & \textbf{99.5}        & \textbf{91.9}  \\ \bottomrule

\end{tabular}%
}
\label{table:result}
\end{table*}

\subsection{Dataset}
We evaluated our model using the ALFRED \cite{ALFRED} benchmark, a comprehensive dataset designed to assess EIF. It demands a set of abilities, including comprehending natural language instructions, planning a sequence of actions to attain the goal, and navigating to the intended location, thus making it a notably challenging task.
The ALFRED dataset comprises 8,055 expert demonstration episodes and is divided into training, validation, and test datasets. The validation and test datasets can be categorized into two distinct groups: "seen" datasets, referring to those whose scenes have been part of the training dataset, and "unseen" datasets, denoting those whose scenes have not been included in the training dataset. Different scenes refer to variations in the types of objects, their shapes, colors, and their relative positions. The ALFRED dataset encompasses a total of 120 distinct scenes, posing a significant challenge for the agents to adapt to diverse environments.

Within the ALFRED \cite{ALFRED} dataset, there are a total of 25,743 language instructions. Multiple language instructions form a set to describe a single expert demonstration episode. To illustrate, in a task like cooling a plate and placing it at a receptacle, the corresponding instructions might include phrases like ``\textit{Put a chilled plate on the counter left of the sink}'' or ``\textit{Grab the plate from the corner, cool the plate in the refrigerator, put the plate by the sink}''.

\subsection{Baseline Methods}
We compared HAPFI with three distinct models: FILM \cite{FILM}, CAPEAM \cite{CAPEAM}, and E.T. \cite{Episodic_transformer}.
\begin{itemize}
\item \textbf{FILM}: 
FILM \cite{FILM}, which utilizes a template-based approach, is widely regarded as a representative model within the ALFRED \cite{ALFRED} benchmark.
It categorizes instructions into seven distinct templates and predicts the corresponding actions, objects, and receptacles for the template predicted to be appropriate for the task. For instance, in the task of cleaning and placing an object, the template is structured as follows: [(\textit{Obj}, PickUp), (SinkBasin, Put), (Faucet, ToggleOn), (Faucet, ToggleOff), (\textit{Obj}, PickUp), (\textit{Recep}, Put)], with the italicized components representing the predicted elements. FILM relies solely on high-level instructions as its data input without utilizing other modalities or historical information.
\item \textbf{CAPEAM}: 
Recognized as the current or state-of-the-art model on ALFRED \cite{ALFRED}, CAPEAM \cite{CAPEAM} harnesses an LSTM-based technique for sub-goal planning. 
The output of CAPEAM is a sequence of subgoals that includes actions, objects, and receptacles. This model exclusively utilizes high-level language instructions without incorporating other modalities or historical information. 
\item \textbf{E.T}.: Differing from the previous two baselines, E.T. incorporates both historical data of images and sub-goals. However, it requires more detailed, step-by-step directives such as ``\textit{Turn right and walk to the dining table}'', instead of the high-level instruction like ``\textit{Put a chilled apple in the microwave}''.

\end{itemize}
\subsection{Results}

The evaluation results are presented in Table \ref{table:result}. Evaluation metric is the percentage of cases where the predicted action, object, and receptacle match the ground truth, relative to the total number of data points. The table clearly illustrates that HAPFI, with all modalities and histories, outperforms other models in action planning of the ALFRED \cite{ALFRED} benchmark. 

The disparity between HAPFI and both FILM \cite{FILM} and CAPEAM \cite{CAPEAM} is rooted in the fact that HAPFI operates as a step-by-step model, setting it apart from the others. This distinctive approach enables HAPFI to consider both its previous actions and prior visual observations, leading to more effective action planning. 
Furthermore, while FILM and CAPEAM rely solely on high-level language instructions, HAPFI incorporates visual inputs alongside corresponding bounding boxes, making its planning process environment-aware. This is advantageous in action planning, as it allows the model to discern valuable relationships between executed actions and objects, as well as their respective locations.

On the other hand, the distinction between E.T.\cite{Episodic_transformer} and HAPFI can be attributed to the method of modality integration. 
While E.T. employs a single multi-modal transformer encoder focusing on uni-modal output, our approach utilizes a deep fusion based methodology, involving two fusion stages and harnessing outputs from both visual and linguistic features. It's worth noting that although E.T. incorporates an attention mechanism within its multi-modal transformer, performing self-attention within mixed modality settings may not effectively facilitate mutual attention, unlike our approach which utilizes cross-attention between distinct modalities. The disparity in performance between E.T. and HAPFI highlights the substantial impact of our early and deep fusion technique in enhancing action planning. It is also important to note that E.T. incorporates step-by-step instructions, which is generally less complex than high-level instructions.

\begin{figure*}[t]
\centering
\includegraphics[width=16cm]{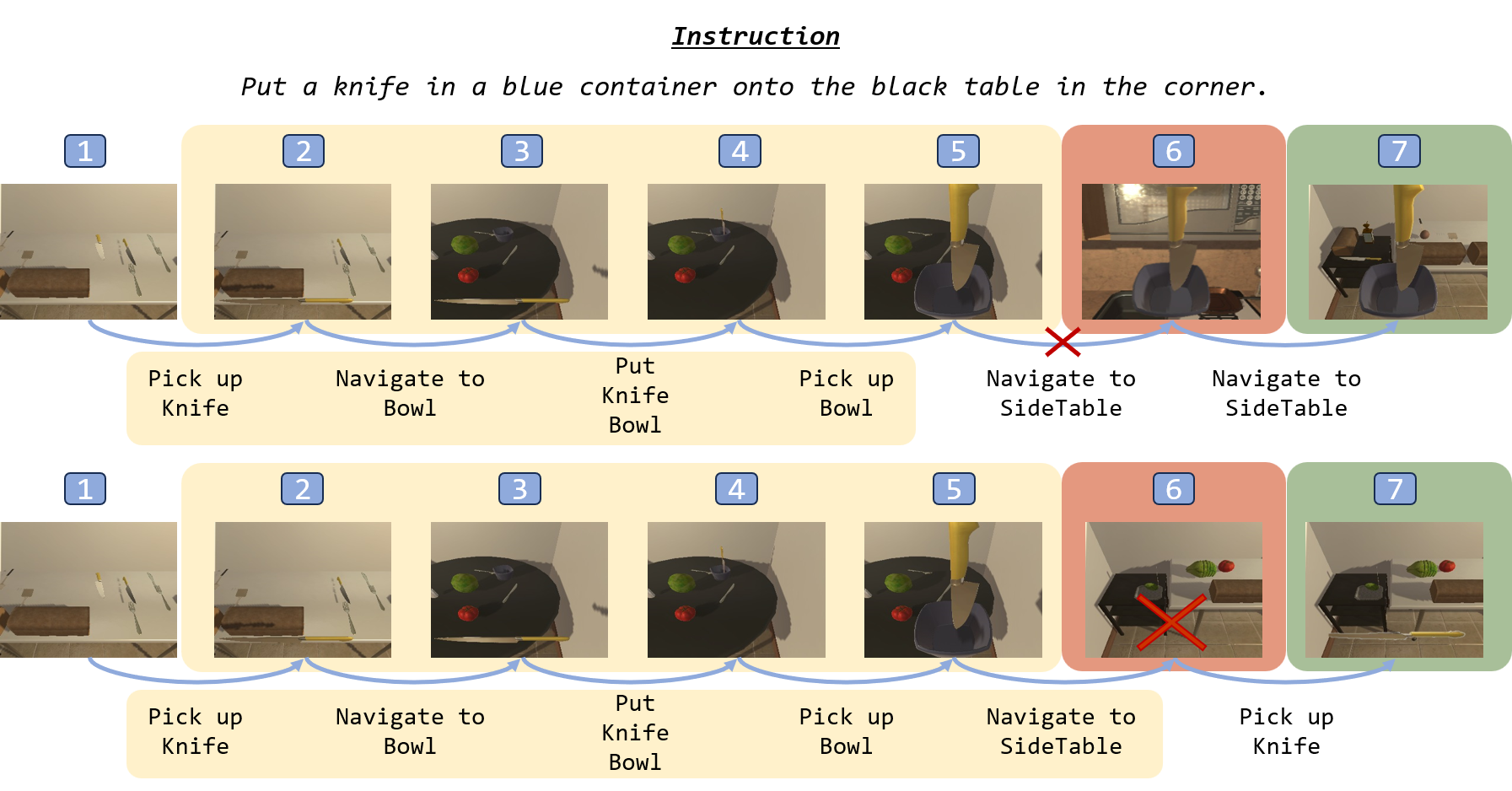}
\caption{\textbf{Illustration of re-planning procedure} This figure illustrates the re-planning procedure. In both rows, the agent performs four actions following the high-level instruction: picking up the knife, navigating to the bowl, putting the knife in the bowl, and picking up the bowl with the knife. In the first row, an erroneous navigation occurs, leading the agent to the microwave. In the second row, a manipulation error results in the agent dropping the bowl and knife.}
\label{fig:qualitive}
\end{figure*}

\noindent \textbf{Ablation Study.}
We conducted experiments involving HAPFI that lacked specific modalities or histories. The modalities and histories employed in each model, along with the respective results, are indicated in Table \ref{table:result}. The results demonstrated that HAPFI, equipped with all modalities and histories, exhibits superior performance in comparison. 
We will proceed to provide an explanation of the results, starting from the first result from the table and moving downward, comparing with the result of the HAPFI with all modalities and histories.

To delve into the details, the performance of the model lacking visual information is inferior to that of the model with all modalities and histories. This is primarily because the model, without access to visual information, remains unaware of the current environmental context, thereby increasing the likelihood of failure. This highlights the critical role played by our multi-modality approach in enhancing the action planning process.

It is evident that a model without visual and sub-goal history experienced a substantial decrease in accuracy. Notably, the decrease in accuracy for validation of unseen data was more pronounced when compared to validation seen data, contrasting with the model's performance that benefits from the utilization of both history types. This implies that incorporating history into the model enhances its robustness against overfitting to the training data. The absence of history confines the agent's perception to only the current scene, which can significantly differ from the training data, particularly in the case of validation unseen data. This incongruity can lead to model confusion due to the unfamiliarity. However, the inclusion of history allows the model to reference past actions, enabling it to flexibly determine the next steps and proceed without failures since the action procedures tend to have a generalized context.

Additionally, it is apparent that utilizing bounding box information yields an advantage in performance. This enhancement stems from the model's increased awareness of the objects within the agent's view and their relative locations, thereby enhancing its overall understanding of the current environment.

\subsection{Qualitative Analysis.} 
Through qualitative analysis, we show that the model is capable of understanding the nature of failures and, consequently, executing appropriate re-planning in response to these failures. An illustrative example of re-planning is provided in Figure \ref{fig:qualitive}. In this specific scenario, the given instruction entails placing a knife in a bowl and relocating it to a side table. After completing step 5, the expected course of action is to navigate to the side table with the knife in a bowl. We considered two potential failure scenarios here.

In one scenario, a navigation error led the agent to the wrong location, e.g., in front of the microwave in Figure~\ref{fig:qualitive}. Upon examining Figure \ref{fig:qualitive}, it becomes evident that the agent has adeptly re-planned its course, redirecting to the side table after ending up at the microwave. 
In the other scenario, a manipulation error resulted in the agent dropping the bowl with the knife, causing the bowl and knife to be out of the agent's view. Figure 2 illustrates how the agent addresses this failure by picking up the knife again after dropping it.

Our experiments have confirmed the model's ability to comprehend the procedural context it has followed.
Without the model's awareness of its history and the current context, the model provides identical plans for both scenarios, lacking the capability to discern how and where it went wrong.

In summary, our experiments not only validate the model's aptitude for accurate performance within the correct context but also highlight its capacity to address failures through re-planning and adaptive problem-solving. This implies that our model is also applicable in real-world scenarios where the behavior of the agent is not perfect.

%%%%%%%%%%%%%%%%%%%%%%%%%%%%%%%%%%%%%%%%%%%%%%%%%%%%%%%%%%%%%%%%%%%%%%%%%%%%%%%%
\section{CONCLUSION}

Throughout this paper, we delved into the challenge of Embodied Instruction Following (EIF). 
In response, we introduced HAPFI, a framework that leverages various types of historical data to enhance the quality of planning. This approach uniquely integrates multiple sources of modalities, creating a more informed agent. 
By doing so, HAPFI not only demonstrates the potential for improved planning capabilities but also underscores its capacity to identify agent failures and rectify them through suitable re-planning strategies. This paves the way for more cognitively adept and adaptable robotic systems, poised to navigate and interact within complex real-world environments with heightened proficiency and understanding.

\section*{Acknowledgements}
This work was partly supported by the IITP (2021-0-02068-AIHub/15\%, 2021-0-01343-GSAI/20\%, 2022-0-00951-LBA/25\%, 2022-0-00953-PICA/25\%) and NRF (RS-2023-00274280/15\%) grant funded by the Korean government.

\bibliographystyle{IEEEtran.bst}
\bibliography{ref}

\end{document}